\documentclass[sigconf]{acmart}

\AtBeginDocument{%
  \providecommand\BibTeX{{%
    \normalfont B\kern-0.5em{\scshape i\kern-0.25em b}\kern-0.8em\TeX}}}

\setcopyright{acmcopyright}
\copyrightyear{2018}
\acmYear{2018}
\acmDOI{XXXXXXX.XXXXXXX}

\usepackage{natbib}
\usepackage{graphicx}
\usepackage{subfigure}
\usepackage{textcomp}
\usepackage{xcolor}
\usepackage{algpseudocode}
\usepackage{algorithm}
\theoremstyle{plain}
\newtheorem{theorem}{Theorem}

\theoremstyle{definition}
\newtheorem{definition}[theorem]{Definition}

\theoremstyle{remark}

\usepackage{enumitem}
\usepackage{xcolor}
\usepackage{pifont}
\usepackage{xspace}
\newcommand{\eg}{\emph{e.g.}\xspace}
\newcommand{\ie}{\emph{i.e.}\xspace}

\newcommand{\greencheck}{{\checkmark}}
\newcommand{\xmark}{\ding{55}}%
\newcommand{\redx}{{\xmark}}

\acmConference[Conference acronym 'XX]{Make sure to enter the correct
  conference title from your rights confirmation emai}{June 03--05,
  2018}{Woodstock, NY}
\acmPrice{15.00}
\acmISBN{978-1-4503-XXXX-X/18/06}

\begin{document}

\title{Learning Cooperative Oversubscription for Cloud by Chance-Constrained Multi-Agent Reinforcement Learning}

\author{Junjie Sheng}
\email{jarvis@stu.ecnu.edu.cn}
\affiliation{%
  \institution{East China Normal University	}
  \country{China}
}
\author{Lu Wang}
\email{luwang@stu.ecnu.edu.cn}
\affiliation{%
  \institution{East China Normal University}
  \country{China}
}
\author{Fangkai	Yang}
\email{fangkai.yang@microsoft.com}
\affiliation{%
  \institution{Microsoft Research}
  \country{China}
}
\author{Bo Qiao}
\email{boqiao@microsoft.com}
\affiliation{%
  \institution{Microsoft Research}
  \country{China}
}
\author{Hang Dong}
\email{hangdong@microsoft.com}
\affiliation{%
  \institution{Microsoft Research}
  \country{China}
}
\author{Xiangfeng Wang}
\email{xfwang@sei.ecnu.edu.cn}
\affiliation{%
  \institution{East China Normal University}
  \country{China}
}
\author{Bo Jin}
\email{bjin@cs.ecnu.edu.cn}
\affiliation{%
  \institution{East China Normal University}
  \country{China}
}
\author{Jun Wang}
\email{jwang@cs.ecnu.edu.cn}
\affiliation{%
  \institution{East China Normal University}
  \country{China}
}
\author{Si Qin}
\email{Si.Qin@microsoft.com}
\affiliation{%
  \institution{Microsoft Research}
  \country{China}
}
\author{Saravan Rajmohan}
\email{Saravanakumar.Rajmohan@outlook.com}
\affiliation{%
  \institution{Microsoft 365}
  \country{United States}
}
\author{Qingwei Lin}
\email{qlin@microsoft.com}
\affiliation{%
  \institution{Microsoft Research}
  \country{China}
}
\author{Dongmei Zhang}
\email{dongmeiz@microsoft.com}
\affiliation{%
  \institution{Microsoft Research Asia}
  \country{China}
}
\renewcommand{\shortauthors}{Anonymous, et al.}

\begin{abstract}
Oversubscription is a common practice for improving cloud resource utilization.
It allows the cloud service provider to sell more resources than the physical limit, assuming not all users would fully utilize the resources simultaneously.
However, how to design an oversubscription policy that improves utilization while satisfying the some safety constraints remains an open problem. 
Existing methods and industrial practices are over-conservative, ignoring the coordination of diverse resource usage patterns and probabilistic constraints.
To address these two limitations, this paper formulates the oversubscription for cloud as a chance-constrained optimization problem and propose an effective Chance-Constrained Multi-Agent Reinforcement Learning (C2MARL) method to solve this problem. Specifically, C2MARL reduces the number of constraints by considering their upper bounds and leverages a multi-agent reinforcement learning paradigm to learn a safe and optimal coordination policy.
We evaluate our C2MARL on an internal cloud platform and public cloud datasets. 
Experiments show that our C2MARL outperforms existing methods in improving utilization ($20\%\sim 86\%$) under different levels of safety constraints.
\end{abstract}





\maketitle

\section{Introduction}
\begin{figure}
    \centering
    \includegraphics[width=0.48\textwidth]{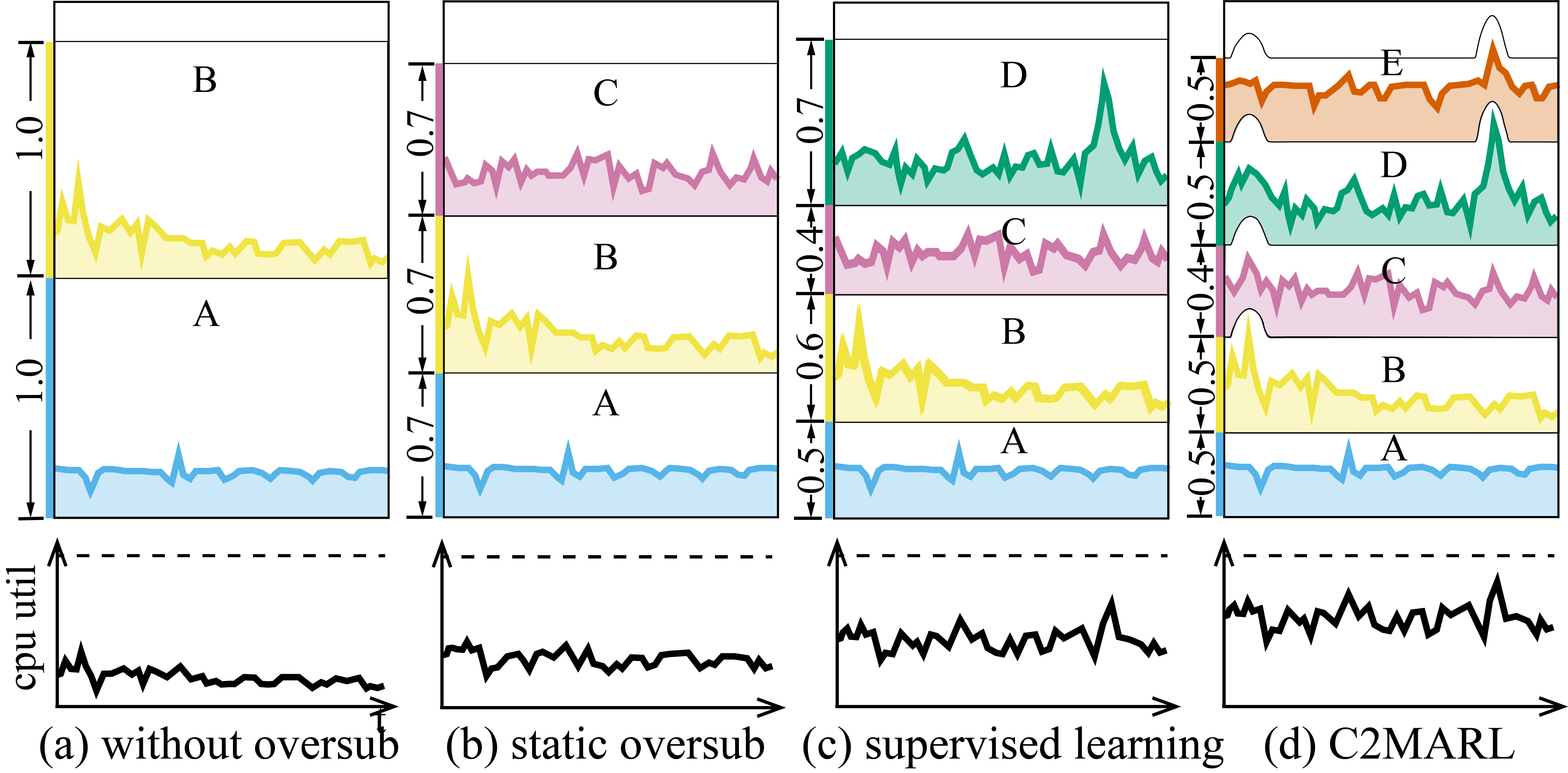}
    \caption{The illustration of different oversubscription strategies. Multiple virtual machines (inner boxes marked with capital letters) are allocated on the same physical machine (outer boxes)  (upper figures). The sum of actual CPU usage of all the VMs (lower figures). The colored area is the actual CPU usage, and the number aside is the oversubscription rate, \ie, assigned resources over requested resources.
    }
    \label{fig:oversub}
\end{figure}

Cloud computing has been widely adopted in recent years. 
Many companies, \eg, Netflix and LinkedIn, request cloud computing resources from cloud providers to host their services and applications~\cite{DBLP:journals/corr/abs-2007-01823}.
The requested resources are usually served by Virtual Machines (VMs), which are virtualized instances hosted on Physical Machines (PMs).
Users (what is called \textit{Subscribers} in this paper) estimate their resource usage and request a number of VMs accordingly. The \textit{actual usage} of computing resources is unstable over time in general (colored areas in Figure~\ref{fig:oversub}), \eg, email applications have peak resource usage during working hours. Then subscribers take the peak usage for references when requesting computing resources, and the \textit{requested resources} are usually overestimated~\cite{predClusterOversub}. In reality, the actual usage does not always peak to reach the requested resources. For example, the services and applications running on the Alibaba cloud platform have less than 50\% actual usage compared to the requested in the majority of time~\cite{guo2019limits}. Therefore, a large amount of requested resources is not utilized, and it leaves an opportunity to reduce such waste and improve resource utilization.


An effective way to reduce the unused resources is  oversubscription~\cite{shlifer1975airline, kumbhare2021prediction}.
For each VM on a PM, oversubscription assigns less than the requested resources to the VM and allows the VM to use beyond the \textit{assigned resources} up to the requested resources when the PM has spare capacity\footnote{Currently cloud providers only do oversubscription for internal users~\cite{amazon}.}.
Figure~\ref{fig:oversub} shows that oversubscription improves resource utilization by allocating more VMs on the same Physical Machine (PM).
Although oversubscription improves resource utilization, it creates risks if it is done inappropriately and overaggressively. The hot machine problem~\cite{hot} is one of the risks that impairs the server reliability. More specifically, if the sum of the actual usage of these VMs exceeds the resource capacity of the host PM, the PM becomes a \textit{hot machine}, and it will have a high risk on crash. Thus, the key challenge is to design an oversubscription policy that reduces the most unused resources while having a low probability of causing hot machines.

\begin{figure}
    \centering
    \includegraphics[width=0.48\textwidth]{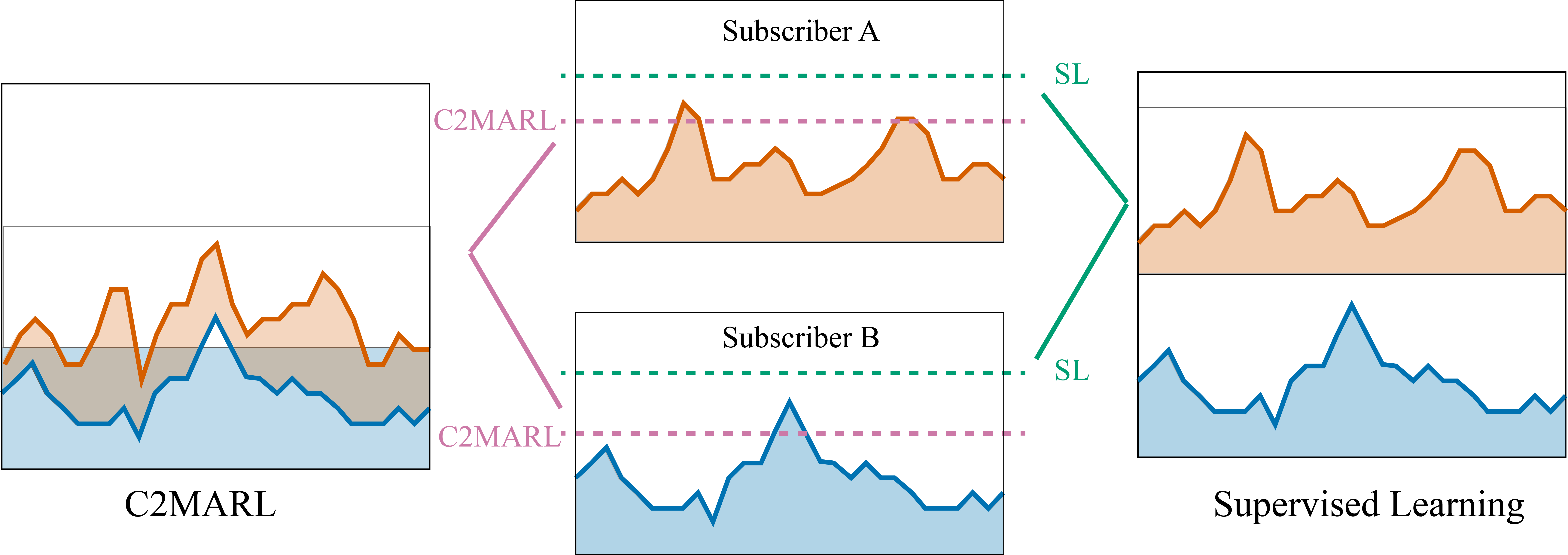}
    \caption{The coordination of two VMs belonging to two subscribers. The supervised learning method makes an over-conservative oversubscription decision (green dashed lines). C2MARL considers staggering the usage peaks. Although the assigned resources (pink dashed lines) is lower than the peak usage for each VM, the sum of usage does not exceed the sum of assigned resources over time, leaving additional capacity for new coming VMs (stacked area plot on the left).}
    \label{fig:marlcoop}
\end{figure}

In industrial practice, gird search~\cite{govindaraju2015big} is a common oversubscription policy. It searches a static oversubscription rate for all subscribers, \ie, computed by assigned resources over requested resources. For example, Figure~\ref{fig:oversub}(b) shows that the grid search oversubscription policy can allocate one more VM to increase resource utilization while avoiding hot machines.
However, the grid search approach is often nonadaptive as it ignores the heterogeneity of actual usage patterns among various subscribers, which limits the utilization improvement. Moreover, it is impractical to conduct the grid search for each subscriber as the search space is growing exponentially with the number of subscribers.
\citet{kumbhare2021prediction} and \citet{chen2018improving} propose supervised-learning-based oversubscription approaches that estimate the actual or the maximal resource usage for each subscriber (Figure~\ref{fig:oversub}(c)). However, existing methods often come up with conservative oversubscription policies for two major reasons: 1) they ignore the coordination of heterogeneous resource usage patterns among subscribers, and 2) they are unable to leverage probabilistic constraints. Figure~\ref{fig:marlcoop} gives an example of the coordination of two subscribers. With the consideration of heterogeneity of usage patterns of different subscribers, \eg, staggered peaks, the assigned resources can be lower than peak usage without causing hot machines. Thus there is a great opportunity with multi-agent coordination to improve resource utilization.
On the other hand, cloud systems are highly stochastic.
The usage of virtual machine is usually hard to predict and there always a small chance that all the virtual machines reach their peak usages simultaneously. 
Avoiding the small chance events inevitably limits a lot on achieving benefits and probabilistic constraints are often taken as good practice for that better describe the constraints~\cite{pan2021constrained}.
The existing methods cannot leverage the probabilistic constraints but rather design oversubscription policies toward fulfilling derterministic safety constraints regardless of their diversity.




To address the above two challenges, this paper formulates the oversubscription problem as a chance-constrained multi-agent optimization problem. In order to coordinate the resource utilization of subscribers under the chance constraints, this paper proposes a novel chance-constrained multi-agent reinforcement learning (C2MARL) method. It brings into value decomposition networks to deal with the unknown dynamics of the multi-agent cloud system, and it proposes a primal-dual reformulation to address the non-convex chance constraints. Overall, C2MARL seeks to maximize resource utilization by saving more resource capacity for the coming VMs while constraining the number of hot machines under the threshold with a high probability (Figure~\ref{fig:oversub}(d) and \ref{fig:marlcoop}).


Our contributions can be summarized as follow: 
\begin{itemize}
    \item This paper formulates the oversubscription problem in Cloud as a chance-constrained optimization problem.
    \item A novel MARL algorithm, C2MARL, is proposed to solve the chance-constrained optimization problem in a large-scale cloud system. To the best of our knowledge, C2MARL is the first method that leverages multi-agent to solve the chance-constrained problem.
    \item Experiments on both an internal cloud platform and public Azure datasets show that C2MARL achieves better performance (improves $20\%\sim 86\%$ in utilization compared with baselines) under different levels of safety constraints.
\end{itemize}

\section{Related Works}
\subsection{Oversubscription in cloud}
In cloud computing, oversubscription is to assign less than requested resources to the VMs on a PM so that more VMs could be allocated on the PM~\cite{householder2014cloud}.
Many resources are considered to be oversubscribed, \eg, CPU~\cite{cohen2019overcommitment}, memory~\cite{predClusterOversub}, power~\cite{kumbhare2021prediction}, etc. This paper only considers CPU/core oversubscription as CPU is one of the most scarce resources in the cloud and the primary resource considered in VM allocation~\cite{protean}. 
The industrial practice applies grid search methods~\cite{govindaraju2015big} to find a global oversubscription rate that maximizes the saved resources, trying not to violate the constraints.
\citet{ghosh2012biting} emphasizes the importance of actual usage patterns when designing oversubscription policies.
\citet{caglar2014ioverbook} and \citet{kumbhare2021prediction} adopt Neural Networks to learn the usage pattern of VMs, and
\citet{chen2018improving} leverages lightweight Brown’s Double Exponential Smoothing to predict usage patterns.
Other oversubscription works propose new resource allocation methods to improve resource utilization~\cite{ cohen2019overcommitment}. In reality, the resource allocator is a complex system and consists of numerous rules and preferences to fulfill multiple objectives~\cite{protean}, which limits the feasibility of incorporating these works into the practical resource allocator.

\subsection{Chance-Constrained Reinforcement Learning}
Chance-constrained reinforcement learning has been prevalent in recent years.
Some model-based chance-constrained RL improve the over-conservative policy with efficient evaluations~\citet{peng2021model, peng2022model}. However, it is challenging to apply model-based methods due to the complexity of the cloud system. Other model-free chance-constrained RL are mainly designed based on penalty methods and Lagrangian methods.
\citet{geibel2005risk} first adds a large penalty on the reward when the chance constraint is violated. 
\citet{paternain2019learning} derives a lower bound of the safe probability and reformulates the chance-constrained RL as a primal-dual optimization problem. However, the oversubscription problem in this paper involves a large number of chance constraints and subscribers, which makes it challenging to leverage existing methods.


\section{Preliminary}
\subsection{Chance Constrained Dec-POMDP} 
\sloppy This section firstly introduces the Chance-Constrained Decentralized Partial Observable Markovian Decision Process (C2Dec-POMDP). 
It can be formally described by a tuple $\langle \mathcal{I}, \mathcal{S},\{\mathcal{A}_i\}, \mathcal{P},\{{\Omega}_i\},\mathcal{O}, \mathcal{R}, T, \gamma, \{\mathcal{C}_k\}_{k=1}^K, \{\delta_k\}_{k=1}^K, \alpha \rangle$.
$\mathcal{I}$ is the set of agents, and each agent corresponds to a subscriber in our setting.
$\mathcal{S}$ is the state space.
$\mathcal{A}_i$ is agent $i$'s action space and $\mathcal{A}:=\times_{i \in \mathcal{I}} \mathcal{A}_{i}$ is the joint action space. 
$\mathcal{P}: \mathcal{S} \times \mathcal{A} \rightarrow \mathcal{S}$ is the transition function and $\mathcal{P}(s'|s, a)$ is the probability that the state transits to $s'$ when taking joint action $a$ in state $s$.
$\Omega_i$ is the agent $i$'s observation space and the $\Omega:= \times_{i\in \mathcal{I}} \Omega_i$ is the joint observation space. 
$\mathcal{O}: \mathcal{S}\rightarrow \Omega$ is the observation function and $\mathcal{O}(o|s)$ describes the probability of joint observation $o$ when current state is $s$.
$\mathcal{R}: \mathcal{S}\times \mathcal{A} \rightarrow \mathbb{R}$ is the reward function and $r(s, a)$ is the immediate reward when taking $a$ in state $s$.
$T$ is the time horizon.
$\gamma$ is the discount factor.
$\mathcal{C}_k$ is the $k$-th constraint and $\mathcal{C}_k(s, a)$ is the immediate $k$-th constraint cost. 
$\delta_k$ is the upper bound on the cumulative $k$-th constraint cost.
$\alpha_k$ is the probability that the cumulative constraint cost is less than the $\delta_k$. 
$\alpha$ is the probability that the constraints need to be satisfied.
Each agent $i$ has a policy $\pi_i(a_i|o_i)$ that maps $o_i$ to the local action $a_i$.
The goal of solving the C2Dec-POMDP is to find a policy for each agent that maximizes the cumulative reward while satisfying $K$ chance constraints: 
\begin{equation}
\begin{aligned}
 \max _{\pi} ~ & J(\pi)=\mathbb{E}_{s_0}[\sum_{t=0}^T r(s_t, a_t)]\\
\text { s.t. } ~ & \operatorname{Pr}\left(\frac{1}{T}\sum\nolimits_{t=1}^T \mathcal{C}_k(s_{t}, a_t)<\delta \right) \geq \alpha, \mkern3mu \forall k\\
 & \hspace{-8pt}  s_{t+1} \sim \mathcal{P}(s_t, a_t);\mkern3mu a_{t}^i \sim \pi_i(\cdot|o_{t}^i); \mkern3mu o_{t} = \mathcal{O}(s_{t}), \forall t, i\\
\end{aligned}
\label{eq: ccdecmdp}
\end{equation}

\section{Problem Formulation}
This section first gives a programming formulation for the oversubscription problem as a chance-constrained optimization problem.
Then we bring it to the sequential decision view and reformulate it as a C2Dec-POMDP.

\subsection{Programming Formulation}
This paper only oversubscribes the CPU resources as the CPU is often the most valuable and scarce resource in VM allocation \cite{protean}.
It can also be easily adopted on memory oversubscription and oversubscription on other resources without loss of generalization.

The cluster has $K$ PMs where each PM $k$ has $B$ resource capacity in CPU cores. There are $N$ subscribers, each subscriber $n$ requests $C_{n}^t$ cores and deletes $D_{n}^t$ cores at timestamp $t$. Note that deleted cores are caused by VM leaving.
We use $E_{n,k}^t$ to represent the assigned cores to subscriber $n$ currently allocated onto PM $k$.
The oversubscription action $A_{n}^t$ is taken for subscriber $n$ at timestamp $t$ to give $C_{n}^t\odot A_n^t$ cores, where $\odot$ is the element-wise product. 
Our goal is to maximize the remaining cores over all $K$ PMs in the whole time horizon and constrain the actual core usage $U_{n,k}^t$ not to exceed the hot PM threshold $\beta B$ for more than $\delta$ times. 
\begin{equation}
    \begin{split}
        \max_{A} \quad & \sum_{t=0}^T\Bigl( \sum_{k=1}^K (B -\sum_{n=1}^N E_{n,k}^t) \Bigr) \\
        \text{s.t.} \quad & Pr\Bigl(\frac{1}{T}\sum_{t=0}^T\mathbf{1}(\sum_{n=1}^N U_{n,k}^t \geq \beta B) < \delta\Bigr)\geq \alpha, \forall k\\
        \quad & E_{n,k}^t=f(E_{n,k}^{t-1}, D_{n}^t, C_{n}^t \odot A_n^t),\\   
    \end{split}
    \label{eq: original}
\end{equation}
where $f$ can be Best-Fit, First-Fit, Round-Robin, or other scheduler methods. Given previous existing cores, VMs to be created and deleted, and oversubscription actions, $f$ is responsible for allocating oversubscribed VMs to different PMs, giving the new reserved core status $E_{n,k}^t$.
It also should be noted that the scheduler needs to meet the memory and network bandwidth constraints (only PM with enough memory and network resources can be selected to allocate the VM).
Due to the memory and network constraints are handled by the scheduler itself, Eq.~\eqref{eq: original} does not explicitly describe it for simplicity.

\subsection{Markovian Formulation}
As the oversubscription action should be taken in an online manner, this section reformulates the chance-constrained programming as the C2Dec-POMDP.
Specifically for the oversubscription problem, each subscriber $i$ is denoted as an agent $i$ and its action space $\mathcal{A}_i$ is the oversubscription rate space which ranges from $0$ to $1$.
The state is composed of the information of all PMs and each agent's current requested resources.
The transition probability $\mathcal{P}$ is determined by the scheduler function $f$ as the last constraints in Eq.~\eqref{eq: original}.
Each subscriber's requested resources are scaled based on the oversubscription rate.
The cloud platform adopts its scheduler to handle current requests based on the oversubscribed resources.
Then the state of all PMs is updated and subscribers make new requests, after which new states are transited.
The reward function is the current remaining cores over all PMs.

\begin{definition}[$\alpha-\delta$ PM Safety] A Physical Machine $k$ achieves $\alpha - \delta$ PM safety under the joint policy $\pi$ if $\operatorname{Pr}\left(\frac{1}{T} \sum_{t=1}^T p(s_0, \dots, s_t|\pi) \cdot \mathcal{C}_k(s_{t})<\delta\right) \geq \alpha$.
The $\mathcal{C}_k$ returns $1$ if the PM $k$ is hot\footnote{The PM is hot if its CPU usage exceeds the hot threshold $\beta B$.} else returns $0$.
\end{definition}
The constraints for the C2Dec-POMDP are that all the PMs satisfy the $\alpha-\delta$ PM safety.
The goal of agents for subscribers is to coordinate with each other to maximize cumulative remaining cores (cumulative rewards) without violating the chance constraints. 
\begin{equation}
\begin{aligned}
 \max _{\pi}\quad & J(\pi)=\mathbb{E}_{s_0}[\sum_{t=0}^T\gamma^t r(s_t, a_t)]\\
 \text { s.t. } \quad & \operatorname{Pr}\left(\frac{1}{T}\sum_{t=1}^T \mathcal{C}_k(s_{t})<\delta |\pi \right) \geq \alpha, \quad \forall k\\
 \quad &  s_{t+1}=p(s_t, a_t);\mkern3mu a_{t}^i = \pi^i(\cdot|o_{t}^i);\mkern3mu o_{t} = O(s_t), \mkern3mu \forall t,i\\
\end{aligned}
\label{eq: ccmmdp}
\end{equation}

There are several challenges in solving such a problem: 
1) the number of chance constraints increases linearly with the number of PMs,
2) the transition function is not available to the agents,
3) Eq.~\eqref{eq: ccmmdp} is a non-convex optimization problem.

\section{Proposed Methods: C2MARL}
To address the aforementioned challenges, this section proposes a novel Chance-Constrained Multi-Agent Reinforcement Learning (C2MARL) method.
This section first reformulates the problem as a cluster-level C2Dec-POMDP problem to address the first challenge.
To establish the relationship between policy and chance constraint satisfaction and solve the non-convex optimization, this section further relaxes the problem as a primal-dual multi-agent learning problem.
After that, C2MARL is formally proposed to optimize the primal-dual multi-agent learning problem and learn the maximal benefits and safe oversubscription policy for each subscriber.

\subsection{Cluster-Level Chance-Constrained Reformulation}
Eq.~\eqref{eq: ccmmdp} has $K$ chance constraints and $K$ is usually a large number in practice.
This often makes the optimization difficult~\cite{ijcai2019-629}.
This section transforms Eq.~\eqref{eq: ccmmdp} to a cluster-level chance constraint optimization problem that has only one constraint, where the \textit{cluster} is the aggregated representation of all PMs.
\begin{definition}[$\alpha-\delta$ Cluster Safety]
A cluster has $\alpha-\delta$ safety under joint policy $\pi$ if $\operatorname{Pr}\left(\frac{1}{T}\sum_{t=1}^T \mathcal{C}_c(s_{t}) < \delta | \pi \right) \geq \alpha$, where $\mathcal{C}_c(s_t):= \max_k \mathcal{C}_k(s_t)$.
\end{definition}

\begin{theorem}
\label{th: cluster}
For any policy $\pi$ that obtain $\alpha-\delta$ cluster safety, it has $\alpha-\delta$ PM safety on all the PMs.
\end{theorem}
A short proof sketch: $\mathcal{C}_c(s_{t})\geq \mathcal{C}_k(s_t)$ for any $t$ and $k$, 
then $ \operatorname{Pr}\left(\frac{1}{T}\sum_{t=1}^T \mathcal{C}_k(s_{t}) < \delta | \pi \right) \geq \operatorname{Pr}\left(\sum_{t=1}^T \mathcal{C}_c(s_{t}) < \delta |\pi \right)$ for any PM $k$.

With the Theorem~\ref{th: cluster}, the $K$ PM constraints can be transformed into one cluster constraint:
\begin{equation}
\begin{aligned}
\operatorname{Pr}\left(\frac{1}{T}\sum_{t=1}^T  \mathcal{C}_c(s_{t})<\delta |\pi \right) \geq \alpha, 
\end{aligned}
\label{eq: ccmmdp2}
\end{equation}
When the Eq.~\ref{eq: ccmmdp2} is satisfied, the original $K$ constraints get satisfied. 
\subsection{Primal-Dual Reformulation}
The Cluster-Level Chance-Constrained Dec-POMDP is still hard to solve due to the unavailable transition function and the non-convex nature of the chance constraint.
Inspired by the \citet{paternain2019learning}, this paper propose a primal-dual reformulation for the problem.
First, the unavailable transition function makes it hard to optimize $\pi_\theta$ with respect to the hot cluster constraint.
To ease the difficulty on estimating hot cluster constraint, this section transforms the hot cluster constraint as follow:
\begin{equation}
\begin{aligned}
    \max_\theta \quad & V(\theta):=\mathbb{E}[\sum_{t=0}^T  r(s_t, a_t)|\pi_\theta] \\
     \text { s.t. } \quad &  U(\theta) < c, U(\theta):= \frac{1}{T}\sum_{t=0}^T  \mathrm{Pr}(\mathcal{C}_c(s_{t})|\pi_\theta)
\end{aligned}
\label{eq: cumm}
\end{equation}
Intuitively, when $c$ is small enough, the feasible policy in Eq.~\eqref{eq: cumm} will avoid hot cluster and further be feasible in Eq.~\eqref{eq: ccmmdp}.
When set $c$ as
$(1-\alpha)\delta$,
the feasible policy in Eq.~\eqref{eq: cumm} achieves $\alpha-\delta$ cluster safety.

\begin{theorem}
For Eq.~\eqref{eq: ccmmdp2} with finite horizon $T$, a joint policy $\pi$ that satisfies the constraint in Eq.~\eqref{eq: cumm} with $c=(1-\alpha)\delta$, the $\pi$ satisfies the constraint in Eq.~\eqref{eq: ccmmdp2}.
\end{theorem}
According to the Reverse Markov Inequality~\cite{levenberg2002reverse},
\begin{equation}
 \operatorname{Pr}(1-\frac{1}{T}\sum_{t=1}^T \mathcal{C}_c(s_t) \leq 1-\delta) \leq  \frac{ \sum_{t=0}^T  \operatorname{Pr}(\mathcal{C}_c(s_{t}))/T}{\delta}   
\end{equation}
The theorem can then be proved accordingly and more details are provided at Appendix.~\ref{app: proof}.

Eq.~\ref{eq: cumm} is a non-convex constrained optimization problem in general~\cite{ding2020natural}.
A common way to solve the problem is to solve its dual relaxation to obtain an approximation solution.
Taking the Lagrangian method~\cite{bertsekas2014constrained}, the Eq.~\ref{eq: cumm} can be approximately solve as:
\begin{equation}
    \max_\theta \min_\lambda \mathcal{L}(\theta, \lambda):= V(\theta) + \lambda(c-U(\theta)).
    \label{eq: pd}
\end{equation}
where $\lambda\in\mathcal{R}_{+}$ is the  lagrange multiplier. 

\subsection{C2MARL}
This section proposes C2MARL to solve the problem~\eqref{eq: pd} and obtain the optimal safe policy for each subscriber.
The problem~\eqref{eq: pd} can be solved in a primal-dual manner.
For the primal problem, C2MARL encode the constraint part to the reward function.
\begin{equation}
    \begin{aligned}
         \max_\theta \quad & \mathbb{E}[\sum_{t=0}^T \gamma^t r_\lambda(s_t, a_t)] \\
        \text { s.t. } \quad &r_{\lambda}(s,a) = r(s,a)+ \lambda(c-\mathcal{C}_c(s)))
    \end{aligned}
    \label{eq: primal}
\end{equation}
The reward in each step can be decomposed into two parts: the remaining cores on all PMs in the cluster, and the constraint violation penalty in the cluster.
Then the primal problem is transformed as typical multi-agent reinforcement learning (MARL) problem and many MARL methods can be adopted to obtain oversubscription policy for each subscriber.
C2MARL takes a simple and powerful multi-agent reinforcement learning method, VDN~\cite{vdn}, as the backbone.
Specifically,  C2MARL estimates the cumulative rewards after taking joint action $a_t$ in $s_t$ through a state-action value network $Q(s_t, a_t;\theta)$.
The $Q(s_t, a_t;\theta)$ is decomposed by $N$ local observation-action value network ($Q^i(o_t^i; \theta^i)$) and a state-value function ($Q^c(s_t^c; \theta^c)$). 
The intuition behind this is that the cumulative reward is influenced by both the current cluster status and each agent's action selection.
Besides that, only the agent (subscriber) that has requests in current state, its current action contributes benefits/costs to the cumulative rewards. 
Thus C2MARL estimates $Q(s_t, a_t;\theta)$ as:
\begin{equation}
    \begin{aligned}
    & Q(s_t, a_t;\theta) = Q^c(s_t^c; \theta^c) + \sum_i^N m_t^i Q^i(o_t^i; \theta^i),   \\    
    \end{aligned}
\end{equation}
where $o_t^i$ is the local observation of agent $i$, and $\theta^c$ and $\theta_i$ are the neural network parameters of cluster value function and subscriber $i$'s local observation-action-value function, respectively.
With the decomposition, the primal problem can be solved accordingly.
Specifically, C2MARL follows the common target network design and the double Q learning to stabilize the training.
Denote the network parameters as $\theta: = [\theta^c, \theta^1, \dots, \theta^N]$.
Define the target parameters $\bar{\theta}: = [\bar{\theta}^c, \bar{\theta}^1, \dots, \bar{\theta}^N]$ where $\bar{\theta}^c$ is the cluster value function parameter  and $\bar{\theta}^i$ is the agent $i$'s target state action value function's parameter.
Then the loss function of the primal problem is: 
\begin{equation}
    \begin{aligned}
          L = & \arg\min_{\theta} ||Q^c(s_t^c; \theta^c) + \sum_{i=1}^N m_t^i Q^i(o^i_t; \theta^i) -(r_\lambda + y )||_2 \\
          \text { s.t. } \quad &  y = Q^c(s_{t+1}^c; \bar{\theta}^c) + \\ & \sum_{i=1}^N Q^i(\max_{a}Q^i(a, o^i_{t+1}; \theta^i), o_{t+1}; \bar{\theta}^i)\\
    \end{aligned}
    \label{eq: bellman}
\end{equation}
After updating the network parameters $\theta$, the target network parameters are soft updated by $\bar{\theta} = (1-\tau) \theta + \tau \bar{\theta}$, where the $\tau$ is the soft update factor.

The dual problem is a convex problem and can be solved with the closed-form solution.
Define the dual problem learning rate as $\eta_\lambda$, 
\begin{equation}
    \lambda = \max(0,\lambda -\eta_\lambda(c-U(\theta^{k+1})))
    \label{eq: dual}
\end{equation}

The details of the algorithm are summarized in  Alg~\ref{alg: pd-adavdn}.
Each agent selects its action with $\epsilon$-greedy exploration based on its individual $Q$ network.
During optimization period, the primal parameters, target parameters and dual parameters are updated sequentially. 
\begin{algorithm}[htb!]
\caption{{\bf{C2MARL}}: Chance-Constrained Multi-Agnet Reinforcement Learning}
\label{alg: pd-adavdn}
\begin{algorithmic}[1]
\State \textbf{Initialization:} the number of over-subscribers $N$, the cluster value network parameters $\theta^c$, each over-subscriber $i$'s state action value network parameters $\theta^i$, discount factor $\gamma$, target network parameters $\bar{\theta}$, dual learning rate $\eta_\lambda$, safety threshold $\alpha$, hot cluster indicator $h'$.
\For{${\hbox{Episode}} = 1, \cdots, M$}
    \State Reset state and obtain the initial state $s_1:= [s_1^c, o_1^1, \dots, o_1^N]$
    \For{$t=1,\cdots,T$ and $s_t \neq$ terminal}
        \For{each agent $i$}
            \State With probability $\epsilon$ pick a random action $a_t^i$ else choose action with the largest value in $Q_t^i(a_t^i, o_t^i)$);
        \EndFor
        \State Execute global actions and get global reward $r_t$, constraint value $c_t$, next state $s_{t+1}:=[s_{t+1}^c, o_{t+1}^1, \dots, o_{t+1}^N]$;
        \State Store $(s_t, a_t, r_t, c_t, s_{t+1})$ to $\mathcal{R}$;
    \EndFor
    \For{$k = 1, \cdots, K$}
        \State Sample a random mini-batch transitions from $\mathcal{R}$;
        \State Update $\theta$ by minimizing Eq.~$10$;
        \State Update target Q network by soft update manner;
        \State Update dual parameter by Eq.~$11$;
    \EndFor
\EndFor
\end{algorithmic}
\end{algorithm}


\section{Experiments}
This section first proposes a practical oversubscription environment, Oversub Gym, where we can access the approximate state transition by replaying the log data.
Then it introduces the baselines and metrics for evaluating our method.
After that, we compare our method with baselines in different scenarios.

\subsection{Dataset}
This paper takes the internal cloud dataset as one of the backbones of the simulator.
It will be released after dropping sensitive information. 
And another dataset (Azure\footnote{\url{https://github.com/Azure/AzurePublicDataset}}) is a public and open-sourced dataset.
There are $13$ subscribers in the internal cloud, while $9$  subscribers are in Azure's cluster.

\subsection{Environment Settings}
This section presents the design of the Oversub Gym.
It contains two environments (internal and public cloud) that have different user patterns~\cite{azsc}. 
The Oversub Gym takes the traces from an internal cloud computing cluster in February 2022 (denoted as Internal Cloud) and traces from a public cloud (Azure\cite{azure}) in 2019 as the data backbone of two environments.
Besides that, the Oversub Gym has two modes: cold-start and ward-start, to approximate real industrial scenarios.
For the cold-start mode, the environment starts from an empty cluster that has no VMs placed in it. For the warm-start mode, the environment starts from the current cluster status with VMs placed as the log-data shows. 
The critical designs of the Oversub Gym as shown below.


\begin{itemize}
    \item \textbf{Reset}. For the Cold-start scenario, the Oversub Gym reset the environment with a sufficient number of PMs without VMs running on them. 
For the Warm-start scenario, it allocates the current running VMs on the corresponding PMs according to the log data. 
\item \textbf{State.}
The Oversub Gym takes the current cluster status, the current requests, and the hour of the day as the state.
For the cluster status, we take a coarse representation: $[\mathit{cpu}^\mathit{allo}, \mathit{cpu}^\mathit{requested}, \mathit{mem}^\mathit{allo}, \mathit{network}^\mathit{allo}]$. Each component is a $N$ dimension vector, and each dimension represents each subscriber's status.
The request status is similarly defined as $[\mathit{cpu}^\mathit{request}, \mathit{mem}^\mathit{request}, \mathit{network}^\mathit{request}]$.
\item \textbf{Transition.} 
The transition model is constructed by the real-world dataset and the allocator. 
Due to the intractable online training in the real-world cloud, we take the log data to simulate the user requests and usage patterns. 
For the allocator, best-fit is the major allocation policy in industrial allocators~\cite{protean}, and we use best-fit to simulate the cloud allocator decision.
\item \textbf{Action.}
Following the industry practice, we consider discrete action space where subscribers can choose over-subscription rate from $[0.2, 0.3, 0.4, 0.5, 0.6, 1]$. 
\item \textbf{Reward.}
The reward is the number of cores saved by oversubscription $(\mathit{cpu}^\mathit{request}-\mathit{cpu}^\mathit{assigned})/Z$ where $Z$ is a large constant for normalizing the reward.
\item \textbf{Constraint.}
The constraint is defined as Eq.~\eqref{eq: original} which requires each PM not hot over $3$ times ($\delta=\frac{1}{40}$) with more than $\alpha$ probability.
\end{itemize}


\subsection{Environment Analysis}
This section analyzes critical patterns in the cloud and emphasis the opportunity that C2MARL can obtain compared with the baselines.

\begin{figure*}[htb!]
    \centering
	\subfigure[Global CPU Usage]{
		\includegraphics[width=0.28\textwidth]{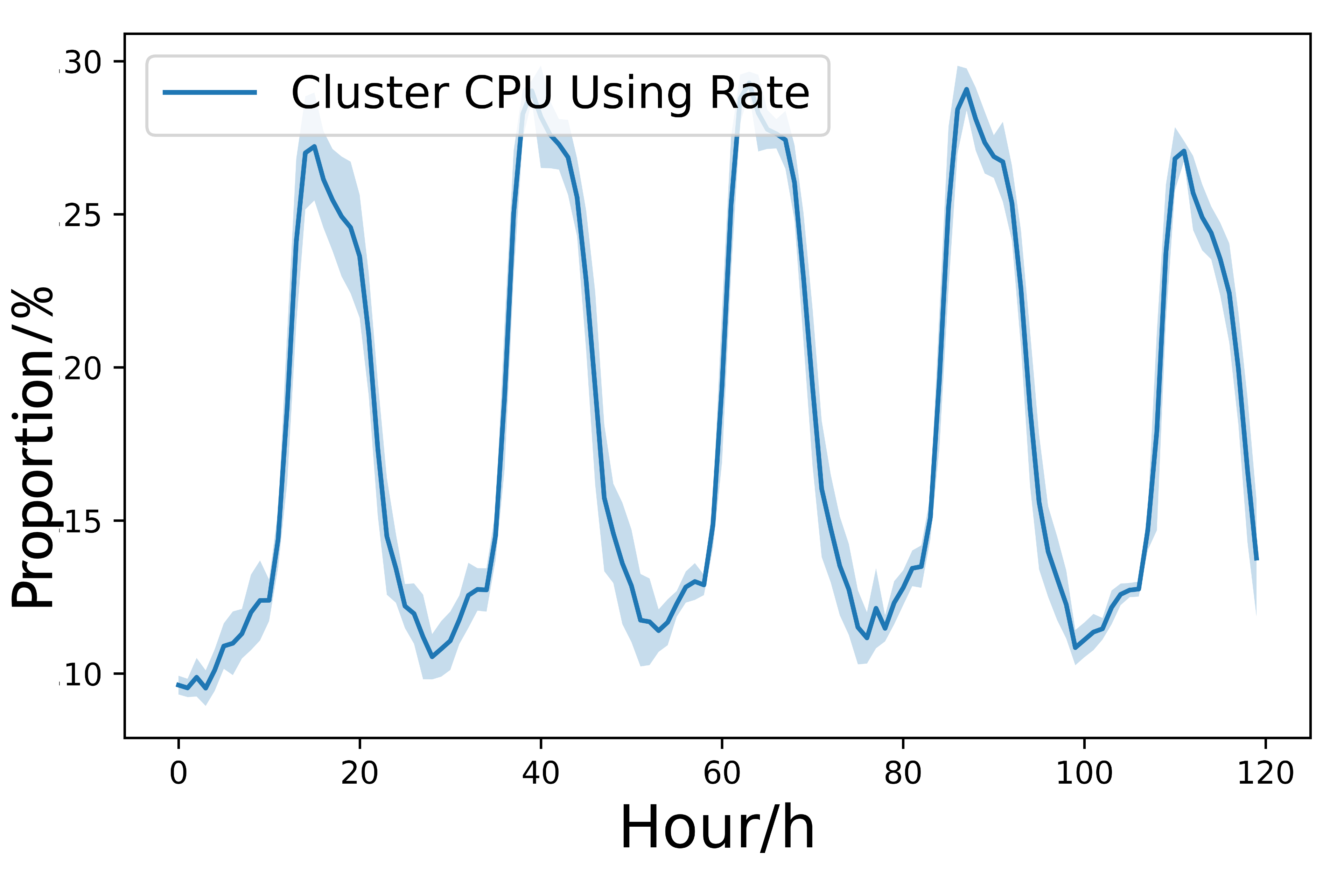}
	\label{fig: cluster-time-cpus}
	}
	\subfigure[Subscriber CPU Usage]{
	\includegraphics[width=0.3\textwidth]{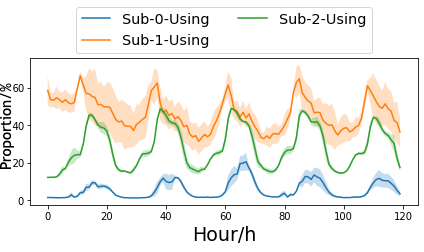}
	\label{fig: sub-rate}
	}
	\subfigure[VM CPU Usage]{
	\includegraphics[width=0.3\textwidth]{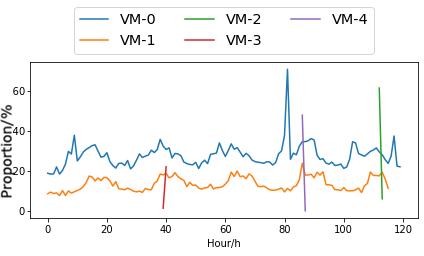}
	\label{fig: vm-rate}
	}
    \caption{CPU usage rate patterns in different levels.
    The \ref{fig: cluster-time-cpus}, \ref{fig: sub-rate} and \ref{fig: vm-rate} show the sampled CPU usage rate in the cluster.
    The cluster's CPU usage has a daily pattern, and the low usage rate indicates the over-estimation nature of the subscribers. Each subscriber has a different CPU usage pattern which brings an opportunity to improve utilization through subscribers' coordination. At the VM-Level, each VM has a different CPU usage peak and duration.}
    
\end{figure*}

\begin{itemize}
    \item \textbf{Cluster-level Pattern.} As shown in the Figure~\ref{fig: cluster-time-cpus}, the CPU usage rate (actual resource usage$/$requested resources) of cluster is range from $0.1$ to $0.3$. 
We observe that nearly $70\%$ requested resources are not used in these $5$ days. Thus there is a great opportunity for oversubscription to reduce the wasted resources.

\item \textbf{Subscription Pattern.}
Figure~\ref{fig: sub-rate} visualizes three subscribers' average usage rate pattern.
We observe that different subscriber has different usage pattern, which motivates us to set diverse oversubscription rate for different subscribers.

\item \textbf{VM Pattern.} Figure~\ref{fig: vm-rate} shows $5$ VMs' usage pattern.
We find that VMs usually have peaks and can be in low-duration. 
The peak usage for a VM is often larger than the VM's average usage rate and it only lasts for a small duration. 
Moreover, the VMs' peaks are often staggered. 
\end{itemize}

\subsection{Metrics and Baselines}
Due to the importance of improving the remaining cores while satisfying the chance constraints, we define metrics to reflect the benefit of the remaining cores and safety. 


The chance-safety is evaluated via \textit{PM-hot ratio} (short as \textbf{PM-Hot-R}),  whether chance-safety ratio $\geq V$ (short as $V$, $V \in \{\textbf{0.75, 0.85, 0.95}\})$.
PM-Hot-R indicates the maximum probability of a PM in the cluster being hot than $\delta$. 

The remaining cores is evaluated via \textit{saved cores} (short as \textbf{S-Cores}). S-Cores is the ratio of the saved cores and total request cores (i.e., $1-$ assigned cores $/$ requested cores). Larger S-Cores indicate better performance.
The saved cores are irrelevant to the initial number of PMs and make a clearer comparison than the remaining cores. 

Three baselines are considered in this paper:
Grid Search~\cite{govindaraju2015big} is an easy-to-implement policy and is widely used in the industry.
It searches for a global and static oversubscription rate for all the subscribers. We consider Grid-0.2, -0.4, and -0.6 as baselines as they are the three best oversubscription rates.
Supervised oversubscription (short as \textbf{SL}) is proposed in the \citet{kumbhare2021prediction} that predicts the maximum usage rate of each subscriber as the subscriber's oversubscription rate.
Moving average (short as \textbf{MA}) oversubscription\cite{chen2018improving} adopts the moving average usage rate for each subscriber as the oversubscription policy.
More experiment details are presented in the Appendix.

\subsection{Internal Cloud (Cold Start)}
\begin{figure}[tb!]
    \centering
    \includegraphics[width=0.42\textwidth]{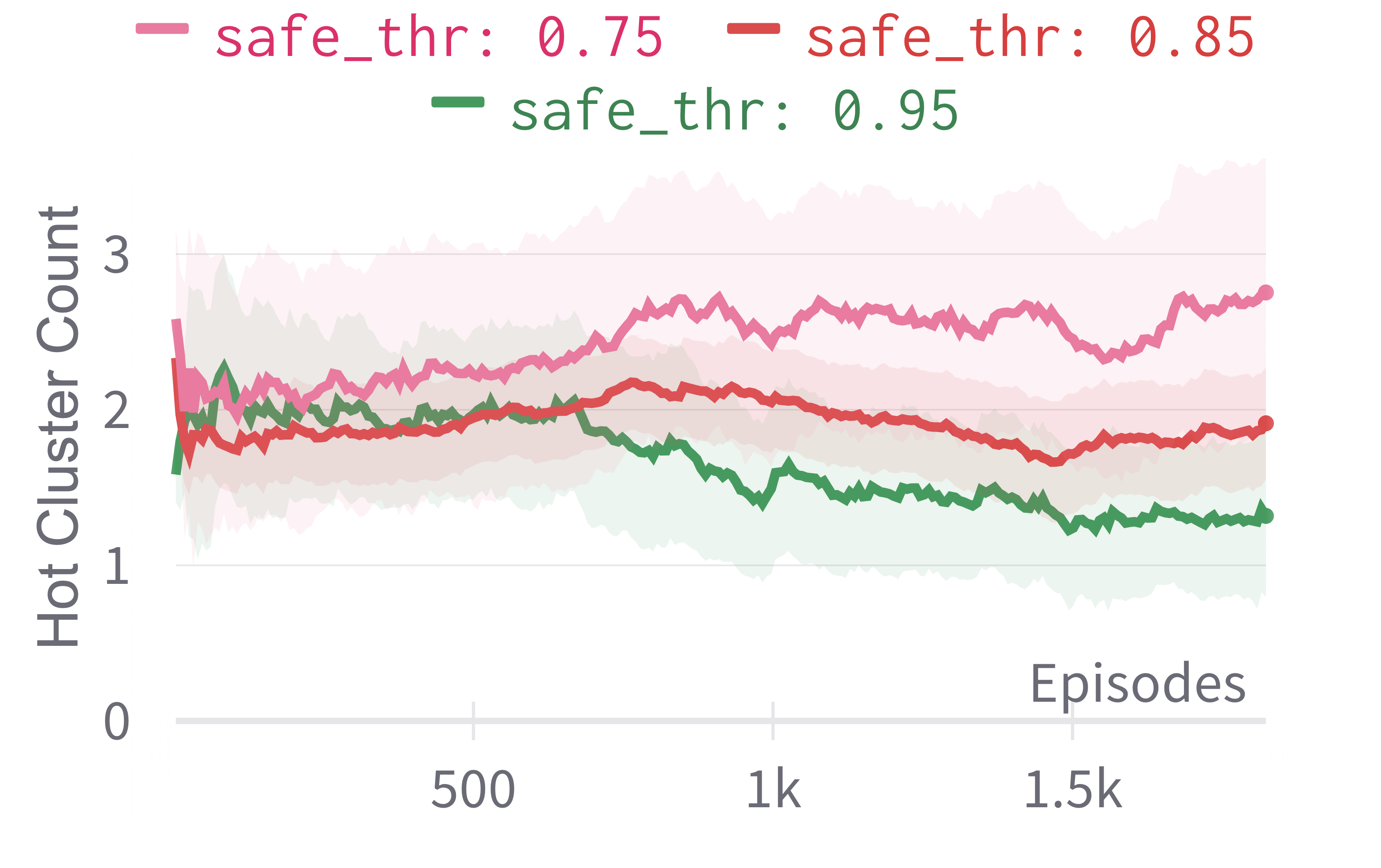}
    \caption{Hot Cluster Count in Internal Cloud with Cold Start. The line is the mean hot cluster count, and the shadow area is the standard deviation.}
    \label{fig:e_hot_cluster}
\end{figure}

\begin{figure}[tb!]
    \centering
    \includegraphics[width=0.42\textwidth]{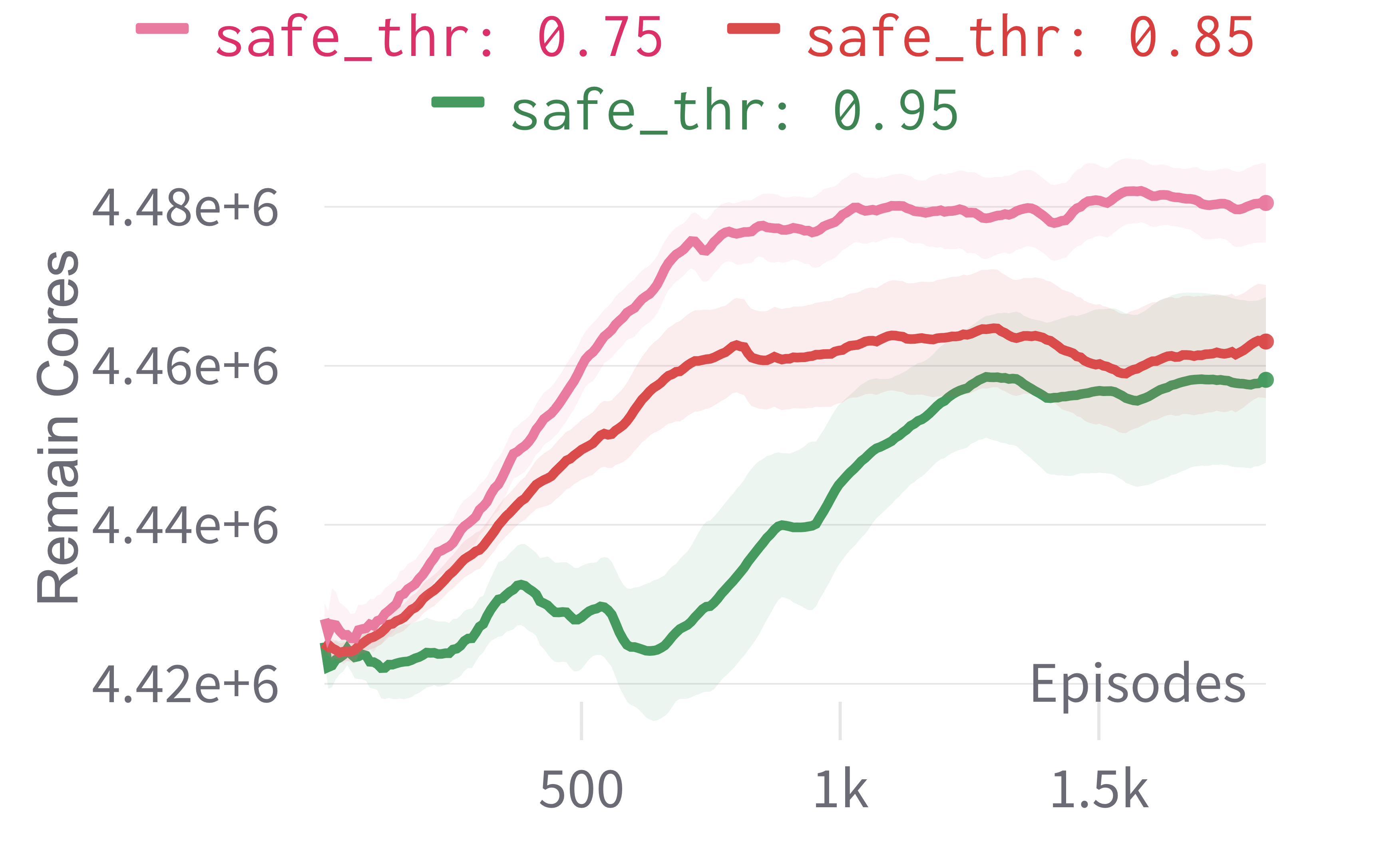}
    \caption{Remain Cores in Internal Cloud with Cold Start. The line is the mean hot cluster count, and the shadow area is the standard deviation.}
    \label{fig:e_remain_cores}
\end{figure}

We first conduct experiments on the cold-start cluster by training our model C2MARL with $1800$ episodes and $3$ different seeds. 
Three different levels of safety preferences are considered $0.75$, $0.85$, and $0.95$.
Figure~\ref{fig:e_remain_cores} shows the remaining cores of C2MARL during training episodes. It can be observes that: (1) with higher safety preference, C2MARL converges to lower remaining cores. 
(2) Figure~\ref{fig:e_hot_cluster} visualizes the hot cluster count during learning, where the higher safety preference leads to less hot cluster count. 

We also test our methods and the baselines in different safety preferences setting with $300$ episodes, where the comparison results are shown in Table~\ref{tb: empty_azsc}. We observe that (1) C2MARL outperforms all the baselines on both chance-safety satisfaction and saved cores with three different safety levels. (2) Grid-0.6, MA, and SL also satisfy all the safety constraints but with sacrificing a large amount of S-cores compared with our C2MARL, which demonstrates that chance constraint can help give more remaining cores.

\begin{table}[htb!]
\caption{Oversubscription in Internal Cloud with Cold Start. The second and third columns are the performance metric with the unit as $\%$. The last three columns are the satisfaction indicator of different levels of safety ($\alpha)$.}
\centering
\resizebox{0.45\textwidth}{!}{

\begin{tabular}{cccccc}
\toprule
Method         & PM-Hot-R & S-Cores &  $0.75$ & $0.85$ & $0.95$   \\ \midrule
Grid-0.2       & $30.9$                              & $80$ & \redx & \redx & \redx          \\ 
Grid-0.4       & $22.7$                             & $60.0$    &     \greencheck & \redx & \redx     \\ 
Grid-0.6       & $0.0$                                    & $40.0$   &      \greencheck & \greencheck & \greencheck     \\ 
MA             & $0$                     & $49.2$      & \greencheck & \greencheck & \greencheck     \\ 
SL             & $0.2\pm0.3$                                 & $49.9\pm3.3$      & \greencheck & \greencheck & \greencheck     \\ 
Our-0.75 & $9.8\pm0.6$     & $66.1\pm6.7$    &    \greencheck & $-$ & $-$  \\ 
Our-0.85 & $6.6\pm0.6$                 & $61.7\pm9.2$ &   \greencheck & \greencheck & $-$ \\ 
Our-0.95 & $3.5\pm 0.5$        & $59.9\pm1.9$     & \greencheck & \greencheck & \greencheck    \\ 
\bottomrule
\end{tabular}
}
\label{tb: empty_azsc}
\end{table}

\begin{figure}
    \centering
    \includegraphics[width=0.45\textwidth]{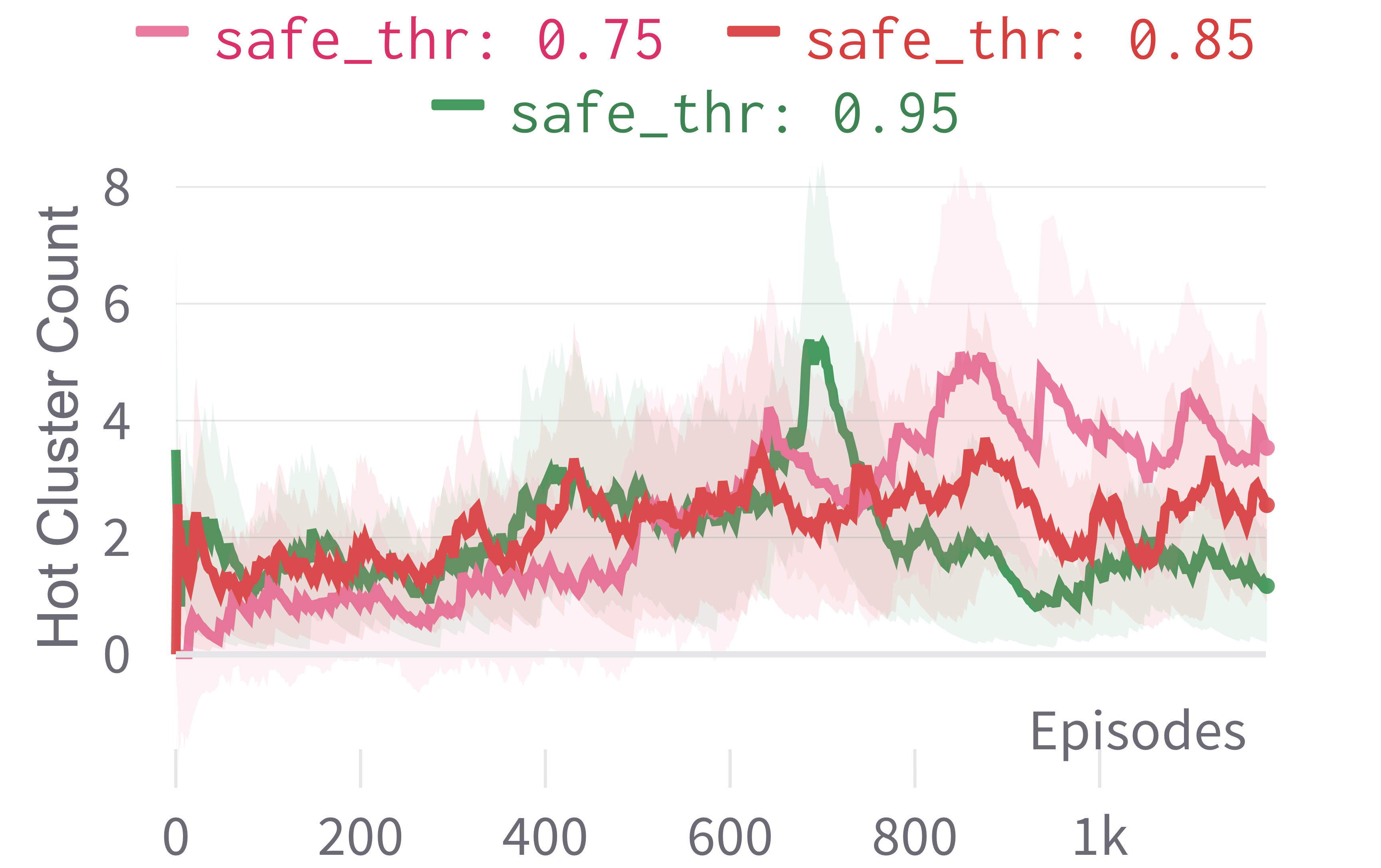}
    \caption{Hot Cluster Count in Internal Cloud with Warm Start. The line is the mean hot cluster count, and the shadow area is the standard deviation.}
    \label{fig:nm_hot_cluster}
\end{figure}
\begin{figure}
    \centering
    \includegraphics[width=0.45\textwidth]{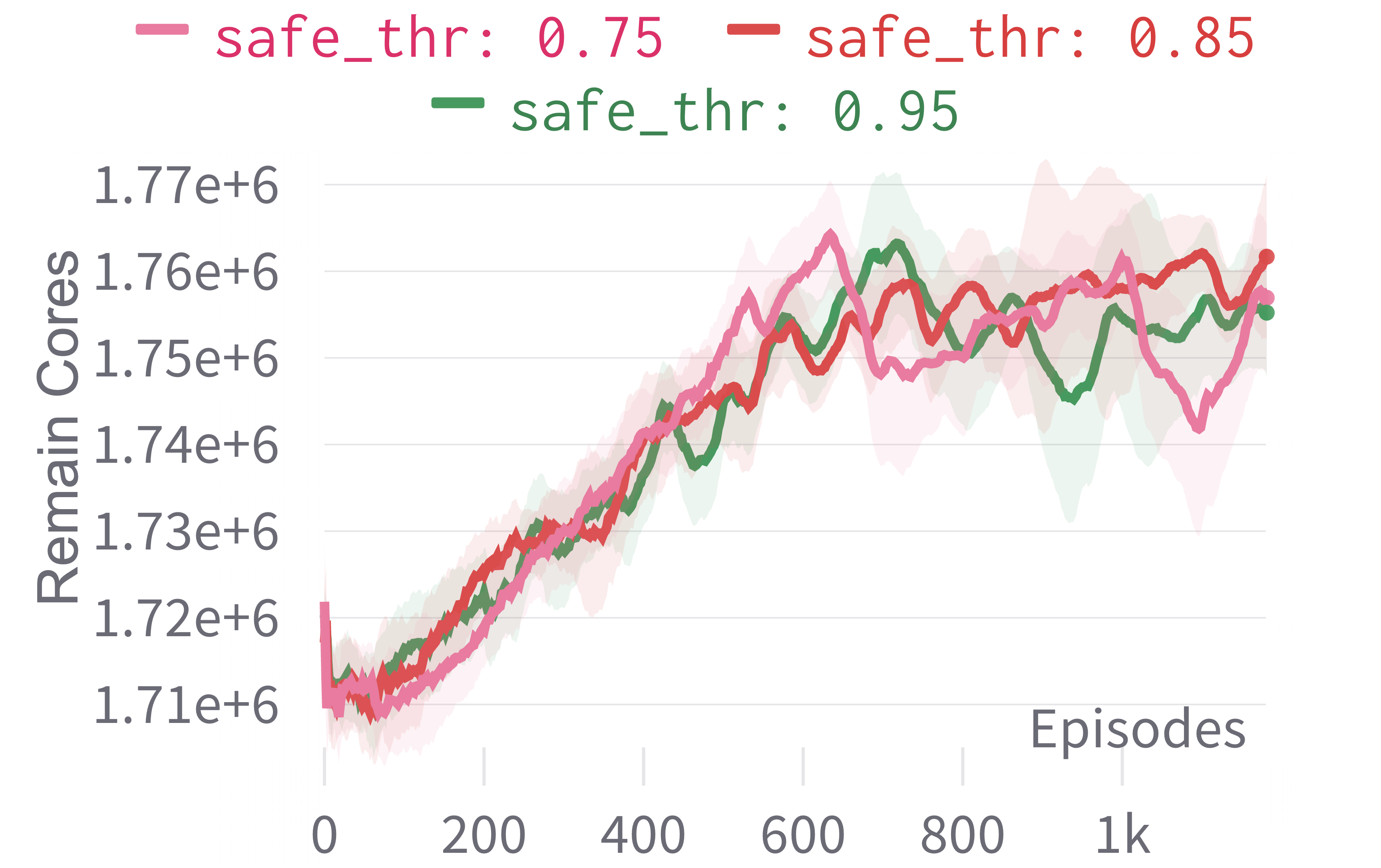}
    \caption{Remain Cores in Internal Cloud with Warm Start. The line is the mean hot cluster count, and the shadow area is the standard deviation.}
    \label{fig:nm_remain_cores}
\end{figure}

\begin{figure}[tb!]
    \centering
    \includegraphics[width=0.42\textwidth]{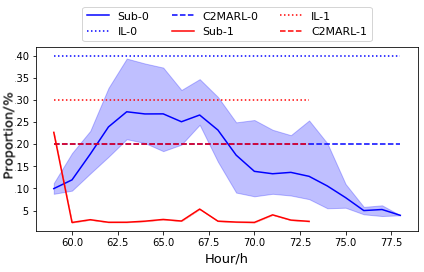}
    \caption{Staggered Peaks. The dashed and dotted are the action of C2MARL and SL agents. The line is the actual usage, and different colors represent different subscribers.}
    \label{fig:case1}
\end{figure}

\begin{figure}[tb!]
    \centering
    \includegraphics[width=0.42\textwidth]{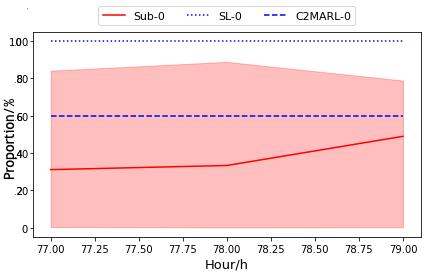}
    \caption{Low Duration. The dashed is the C2MARL's oversubscription action and the dotted is the SL's action. The line is the actual usage.}
    \label{fig:case2}
\end{figure}

\textbf{Case Study.} To analyze how C2MARL-$0.95$ outperforms the baselines, we visualize two randomly selected cases in Figure~\ref{fig:case1} and Figure~\ref{fig:case2}. 
At the time $59$, two subscribers request VMs and their VMs' usage rates peak at different times. 
SL selects an oversubscription rate that is much larger than the peak usage for each subscriber to mimic the largest usage rate.
Our C2MARL takes a more effective action (\ie $0.2$ for the two subscribers) that exploits the staggered peaks via coordination.
At the time $77$, one subscriber requested VMs with a maximum usage larger than $0.8$,
SL takes a conservative action (larger than $0.8$) while our C2MARL takes $0.6$.
This does not make the $0.95$ chance constraints being violated due to the low duration of the VMs ($2$ hours) and large variance of the usage.
By exploiting the opportunity of the staggered peaks and low duration of some VMs, our C2MARL obtains a great performance on the $S$-$Cores$ while satisfying the safety constraints.

\subsection{Internal Cloud (Warm Start)}
In practice, the oversubscription policy is usually applied to non-empty clusters (Warm Start), where a set of VMs has been placed in the cluster in advance. 
We extract these pre-allocated VMs from log data that exist before the oversubscription task to initiate the cluster as a warm-start setting.
We train C2MARL with $1200$ episodes in $3$ different seeds, and the results are shown in Figure~\ref{fig:nm_hot_cluster} and \ref{fig:nm_remain_cores}.
Figure~\ref{fig:nm_hot_cluster} visualizes the hot cluster count during learning, where the higher safety preference leads to a less hot cluster count. Figure~\ref{fig:nm_remain_cores} visualizes the remaining cores during learning, where there is a less clear monotonicity relationship between the safe threshold and the remaining cores. The reason would be that the remaining cores are impacted by the oversubscription actions and the allocation policy. It can be observed that the remaining cores converge to similar points for different safety thresholds in the warm start.
We test the C2MARL and the baselines on the warm-start setting for $300$ episodes, where the results are shown in Table~\ref{tb: non-empty_azsc}.
It can be observed that (1) our C2MARL still outperforms the baselines in different safety-preferences settings. (2) Compared with the cold-start setting (in Table~\ref{tb: empty_azsc}), our C2MARL takes the coordination among pre-allocated VMs as well as the oversubscribed VMs into consideration and obtains more saved cores than the cold-start setting. However, the baselines achieve the same saved cores without any improvement which demonstrates the benefits of coordination.

\begin{table}[tb!]
\centering
\caption{Oversubscription in Internal Cloud with Warm Start. The second and third columns are the performance metric with the unit as $\%$. The last three columns are the satisfaction indicator of different levels of safety($\alpha)$.}
\resizebox{0.45\textwidth}{!}{

\begin{tabular}{cccccc}
\toprule
Method         & PM-Hot-R & S-Cores &  $0.75$ & $0.85$ & $0.95$   \\ \midrule
Grid-0.2       & $95.4$                                 & $80.0$          & \redx & \redx & \redx        \\ 
Grid-0.4       & $100$                                    & $60.0$          & \redx & \redx & \redx        \\ 
Grid-0.6       & $0.0$                                      & $40.0$         & \greencheck & \greencheck & \greencheck         \\ 
MA             & $0.0$                                     & $49.2$        & \greencheck & \greencheck & \greencheck        \\ 
SL             & $0.8\pm1.1$                            & $49.9\pm3.3$              & \greencheck & \greencheck & \greencheck        \\ 
Our-0.75 & $11.6\pm 10.6$               & $69.2\pm 1.5$    & \greencheck & $-$ & $-$  \\ 
Our-0.85 &      $6.8\pm6.8$                             &        $65.1\pm4.8$         & \greencheck & \greencheck & $-$    \\ 
Our-0.95 & $3.5\pm1.5$                 & $69.0\pm 1.4$    & \greencheck & \greencheck & \greencheck   \\ 
\bottomrule
\end{tabular}
}
\label{tb: non-empty_azsc}
\end{table}

\subsection{Azure Cloud}
To verify the effectiveness of C2MARL on different patterns of usage rate patterns, we further evaluate our C2MARL in an open public cloud dataset, Azure.
The average usage rate of Azure is much higher than the internal Cloud.
As shown in Table~\ref{tb: azure-tb}, Grid fails to satisfy the safe constraint due to that most of the users have large usage rates.
When taking no oversubscription (\ie Grid-1.0), the safety constraints are satisfied but no $S$-$Cores$ then.
The MA and SL achieve the over-conservative policies that satisfy the safety constraints.
Our C2MARL obtains a safety satisfaction oversubscription policy under large usage rate datasets while outperforming others on the S-Cores (\ie in 0.95 safety threshold, our-0.95 improves more than $26.8\%$ on S-Cores to the MA and SL). 
\begin{table}[htb!]
\caption{Oversubscription in Azure Cloud. The second and third columns are the performance metric with the unit as $\%$. The last three columns are the satisfaction indicator of different levels of safety($\alpha)$.}
\centering
\resizebox{0.45\textwidth}{!}{

\begin{tabular}{cccccc}
\toprule
Method         & PM-Hot-R & S-Cores &  $0.75$ & $0.85$ & $0.95$   \\ 
\midrule
Grid-0.2       & $100.0$                                 & $80.0$          & \redx & \redx & \redx        \\ 
Grid-0.4       & $100.0$                               & $60.0$ & \redx &\redx & \redx        \\ 
Grid-0.6       & $100.0$                                  & $40.0$         & \redx & \redx & \redx         \\ 
MA                       & $0.0$                & $8.2$         & \greencheck & \greencheck  & \greencheck    \\ 

SL             & $0.0\pm0.0$                                        & $7.6\pm2.2$        & \greencheck & \greencheck & \greencheck        \\ 
Our-0.75 & $19.0\pm 3.8$    & $15.3\pm0.13$    & \greencheck & $-$ & $-$  \\ 
Our-0.85 &      $8.3\pm 3.1$            & $14.6\pm 0.1$                                       & \greencheck & \greencheck & $-$    \\ 
Our-0.95 & $0.0\pm 0.0$                     & $10.4\pm 0.2$    & \greencheck & \greencheck & \greencheck   \\ 
\bottomrule
\end{tabular}
}
\label{tb: azure-tb}
\end{table}
\section{Conclusion}
This paper formulates the oversubscription in Cloud as a chance constraint problem to match the real scene. To solve this problem, we propose an effective chance-constrained multi-agent reinforcement learning method, C2MARL. To reduce the number of chance constraints, C2MARL considers the upper bounds of them and leverages a multi-agent reinforcement learning paradigm to learn a safe and optimal coordination policy. Experimenting on two real-world datasets, C2MARL satisfies all the safety constraints and achieves the largest remaining cores.

\newpage
\bibliographystyle{ACM-Reference-Format}
\bibliography{sample-base}

\appendix
\section{Proofs}
\subsection{Proof of Theorem 4}
\label{app: proof}
We present the detailed proof of theorem 4 here.
For a joint policy $\pi$, if it satisfies the constraint in Eq.~5 with $c=(1-\alpha)\delta$, 
\begin{equation}
    \frac{\sum_{t=0}^T  \mathrm{Pr}(\mathcal{C}_c(s_{t})|\pi_\theta) /T}{\delta} \le 1-\alpha
\end{equation}
With the Reverse Markov Inequality, we have
\begin{equation}
     \operatorname{Pr}(1-\frac{1}{T}\sum_{t=1}^T \mathcal{C}_c(s_t) \leq 1-\delta) \leq  \frac{ \sum_{t=0}^T  \operatorname{Pr}(\mathcal{C}_c(s_{t}))/T}{\delta}  , 
\end{equation}
If 
\begin{equation}
    \frac{1}{T}\sum_{t=0}^T  \mathrm{Pr}(\mathcal{C}_c(s_{t})|\pi_\theta) < (1-\alpha)\delta, 
\end{equation}
then
\begin{equation}
    \operatorname{Pr}(1-\frac{1}{T}\sum_{t=1}^T \mathcal{C}_c(s_t) \leq 1-\delta) < 1-\alpha
\end{equation}
Due to $\operatorname{Pr}(1-\frac{1}{T}\sum_{t=1}^T \mathcal{C}_c(s_t) \leq 1-\delta) = \operatorname{Pr}(\frac{1}{T}\sum_{t=1}^T \mathcal{C}_c(s_t) \geq \delta)$, and $\operatorname{Pr}(\frac{1}{T}\sum_{t=1}^T \mathcal{C}_c(s_t)\geq \delta) + \operatorname{Pr}(\frac{1}{T}\sum_{t=1}^T \mathcal{C}_c(s_t)< \delta)=1$, 
\begin{equation}
    \operatorname{Pr}(\frac{1}{T}\sum_{t=1}^T \mathcal{C}_c(s_t) < \delta) \geq \alpha
\end{equation}
The theorem gets proved. 
\section{Experiment Details}

\subsection{Hyper Parameters}
\begin{table}[htb!]
\footnotesize
\centering
\caption{Hyperparamaters for C2MARL in internal and azure cloud. $a/b$ denote that $a$ is internal cloud's param and $b$ is the azure's.}
\begin{tabular}{c c} 
\hline
number of agents & $12/9$ \\
number of PMs & $500/200$ \\
$\delta$ & $3$ \\
$\beta$ & $0.6$ \\
discount factor & $0.9$ \\
batch size & $10$   \\
memory capacity & $360$ \\
target update factor $\tau$ & $0.001$ \\
Activation function & Relu\\
optimizer & Adam \\
\hline
\end{tabular}
\label{tb:hyperparas}
\end{table}
As for the learning rate and dual learning rate we run grid search from $1e-4$ to $20$ for each setting. 
For all the baselines, we grid search the hyperparameters for their best performances.
All the experiments are trained with requests on $5$ weekdays (from Monday to Friday). 
When it comes to the evaluation, each subscriber's VM CPU usage in a specific hour-in-a-day (the hour modular $24$) is generated from a Gaussian distribution to evaluate the reliability of the obtained policy. 
The Gaussian distribution's mean value and standard deviation are determined by the dataset's statistical mean and standard deviation. 
Suppose the policy obtains high benefits and satisfies the safety constraints in the stochastic usage distribution. In that case, cloud service providers can be more confident in applying it to the real world.


\subsection{Relation between Cluster Hot and PM Hot}
\begin{table}
\caption{Oversubscription in Internal Cloud with Cold Start.}
\begin{tabular}{|l|l|l|l|}
\hline
Method         & PM-Hot-R & C-Hot-R & S-Cores \\ \hline
Grid-0.2       & $30.9$               & $97.2$                  & $80$    \\ \hline
Grid-0.4       & $22.7$               & $72.7 $                 & $60.0$       \\ \hline
Grid-0.6       & $0.0$                  & $0.0$                    & $40.0$   \\ \hline
MA             & $0$                  & $0$                     & $49.2$       \\ \hline
SL             & $0.2\pm0.3$                  & $0.6\pm0.8$                     & $49.9\pm3.3$     \\ \hline
Our-0.75 & $9.8\pm0.6$      & $21.9\pm0.3$         & $66.1\pm6.7$     \\ \hline
Our-0.85 & $6.6\pm0.6$        & $13.9\pm1.2$          & $61.7\pm9.2$  \\ \hline
Our-0.95 & $3.5\pm 0.5$       & $4.7\pm0.1$           & $59.9\pm1.9$    \\ \hline
\end{tabular}
\label{tb: empty_azsc_2}
\end{table}
C-Hot-R indicates the probability of the cluster has hot PM.
Take the Internal Cloud with Cold Start As an example.
As shown in Table~\ref{tb: empty_azsc_2}, the hot cluster rate always higher than the PM-Hot-R and our methods meet the the desired hot cluster safety.

\end{document}